\pdfoutput=1
\documentclass[10pt,twocolumn,letterpaper]{article}

\usepackage[pagenumbers]{cvpr}

\usepackage[T1]{fontenc}
\usepackage{multirow}

\usepackage{microtype}

\newcommand{\secondary}[1]{{\scriptsize\color{gray}#1}}%
\newcommand{\yes}{\checkmark}%
\newcommand{\no}{$\times$}%
\newcommand{\nodata}{-}%
\newcommand{\lb}{$\downarrow$}%
\newcommand{\hb}{$\uparrow$}%

\definecolor{cvprblue}{rgb}{0.21,0.49,0.74}
\usepackage[pagebackref,breaklinks,colorlinks,allcolors=cvprblue]{hyperref}

\def\confName{CVPR}
\def\confYear{2027}

\title{BEVCast: Path-Adaptive BEV for Efficient End-to-End Autonomous Driving}

\author{Qiren Wang, Mingyu Guo, Zixuan Wang, and Weisong Shi\\
Connected and Autonomous Research Laboratory\\
University of Delaware, Newark, USA\\
$\{$qirenw, myuguo, zxwang, weisong$\}$@udel.edu}

\begin{document}
\maketitle

\begin{abstract}
Camera based driving models build a dense bird's eye view (BEV) grid and spend the same computation on every cell of it, although the vehicle crosses only a few of those cells within the planning horizon. We present BEVCast, a two stage BEV whose resolution adapts to the predicted driving path. A coarse pass encodes the whole scene on a grid of 4.096 m cells. A heat head trained on the recorded path assigns every coarse cell a score that we call its heat, and the cells with the highest heat are refined into 1.024 m cells under a fixed budget. The remaining cells keep their coarse resolution, so the trajectory head reads a mixed resolution BEV that still covers the entire scene. BEVCast uses neither an HD map nor a navigation command. Against a full grid of 1.024 m cells with the same architecture and training recipe, BEVCast at a 10\% refinement budget encodes 28\% of the BEV queries and needs 66\% of the BEV encoder time at batch size 1. On nuScenes its mean displacement error is 1.078 m against 1.149 m and its drivable area violation rate is 1.46\% against 2.29\%. On NAVSIM v2 navtest its mean displacement error is 0.577 m against 0.598 m and its extended predictive driver model score is 82.51 against 81.28. Next to published planners BEVCast reaches comparable results on both benchmarks without a navigation command.
\end{abstract}

\section{Introduction}
\label{sec:intro}

Camera based end to end driving integrates visual scene understanding with trajectory planning in one model~\cite{hu2023planning,jiang2023vad}. Many such systems rely on a bird's eye view (BEV) representation, which brings the observations of several cameras into a common spatial frame for perception and prediction and planning~\cite{li2022bevformer,hu2023planning,jiang2023vad}. This spatial organization lets the model reason about the road layout and the surrounding traffic when it predicts the future motion of the ego vehicle. For onboard operation these models must also produce updated trajectories in time with limited computational resources. The cost of building and processing the BEV is therefore central to the balance between prediction quality and the computational demands of driving.

Many BEV based driving models represent the surrounding scene on a dense grid of uniform spatial resolution~\cite{li2022bevformer,hu2023planning,jiang2023vad}. This design couples the local spatial precision to the cost of processing the entire scene at the same resolution, so finer detail around the path of the vehicle also raises the computation everywhere else. Yet on nuScenes the path driven in the next 3 s crosses on average 5 of the 625 cells of 4.096 m that cover the scene. Recent approaches reduce the dense processing through sparse object or instance queries~\cite{liu2023sparsebev,sun2024sparsedrive}, through a perception guided selection of BEV cells~\cite{chambon2024pointbev} or through image features sampled along candidate trajectories~\cite{guo2025ipad}. These designs differ in their selection criterion and in how much of the scene they keep. For ego trajectory prediction the path that the vehicle is about to drive provides a natural spatial prior for concentrating fine detail, while the surrounding road and traffic remain relevant to the driving decision. The challenge is to adapt the BEV resolution to this path while explicit context is retained across the full scene.

To address this challenge we present BEVCast, a path adaptive BEV built in two stages (Figure~\ref{fig:overview}). It combines fine spatial detail along the predicted driving path with coarse coverage of the full scene. A coarse pass first encodes the whole scene and a heat head assigns every coarse cell a heat, a score for whether the vehicle will drive through it within the prediction horizon. The heat head is trained on the recorded path, which every driving log provides without additional annotation. Under a fixed refinement budget the coarse cells with the highest heat are subdivided into fine cells. The main encoder then processes these fine cells together with the unrefined coarse cells and a trajectory head regresses a single trajectory from the resulting mixed resolution BEV. BEVCast takes the camera images and the ego motion derived from consecutive poses as input and uses neither an HD map nor a navigation command.

We evaluate BEVCast on nuScenes~\cite{caesar2020nuscenes} under the open loop protocol of ST-P3~\cite{hu2022stp3} and on NAVSIM v2~\cite{dauner2024navsim,cao2025pseudosim} with the extended predictive driver model score (EPDMS). Controlled comparisons use a full grid of 10{,}000 cells of 1.024 m and a coarse only model without refinement, both with the same backbone and training recipe. The coarse grid alone already retains much of the accuracy of the full grid, and the learned refinement lowers its displacement error by 0.09 m on nuScenes and raises its EPDMS by 0.66 points on NAVSIM. Next to published planners that receive a navigation command BEVCast reaches comparable results on both benchmarks without one.

Our contributions are the following.
\begin{itemize}[leftmargin=*,nosep]
\item We introduce BEVCast, a path adaptive BEV built in two stages for camera based ego trajectory prediction. It combines fine resolution along the predicted driving path with coarse coverage of the full scene and uses neither an HD map nor a navigation command.
\item We design a path supervised refinement that needs no additional annotation. A heat head trained on the recorded path predicts for every coarse cell whether the vehicle will drive through it, and a fixed budget of the cells with the highest heat is subdivided into fine cells.
\item Experiments on nuScenes and NAVSIM show that BEVCast at a 10\% refinement budget encodes 28\% of the BEV queries of a full grid with the same backbone and needs 66\% of its BEV encoder time at batch size 1. Its mean displacement error on nuScenes is 1.078 m against 1.149 m, and its EPDMS on NAVSIM is 82.51 against 81.28.
\end{itemize}

\section{Related Work}
\label{sec:related_work}

\subsection{Scene representations for camera based driving}
\label{sec:related_scene_representations}

BEV representations organize the observations of several cameras in a common spatial frame for perception and ego trajectory prediction. BEVFormer~\cite{li2022bevformer} builds BEV features through spatial cross attention and temporal attention, while UniAD~\cite{hu2023planning} connects this scene encoding with perception and prediction and planning through task specific queries. Subsequent approaches explore different representations for the planner. VAD~\cite{jiang2023vad} represents agents and map elements as vectors and plans through query interactions while it retains a BEV encoder. SparseDrive~\cite{sun2024sparsedrive} learns sparse instance representations for detection and tracking and mapping and couples them with parallel motion prediction and planning. SSR~\cite{li2025ssr} compresses dense BEV features into 16 tokens conditioned on the navigation command. These approaches reshape the scene representation supplied to planning, from spatial BEV features to vectors and instances and compact tokens. Diffusion heads such as DiffusionDrive~\cite{liao2025diffusiondrive} instead replace the regression head by a denoising decoder and are complementary to the representation studied here. On nuScenes these planners receive a navigation command that ST-P3~\cite{hu2022stp3} derives from the recorded future.

\subsection{Sparse and adaptive BEV representations}
\label{sec:related_sparse_adaptive}

Sparse representations reduce spatial processing through different choices of query support. SparseBEV~\cite{liu2023sparsebev} detects objects from several cameras with sparse object queries and builds no dense BEV map. PointBeV~\cite{chambon2024pointbev} instead samples sparse spatial locations for BEV segmentation, and its two pass coarse to fine processing concentrates the sampling around high scoring locations. VectorFormer~\cite{chen2024vectorformer} combines a low resolution BEV with high resolution vector features for detection. Sparsity over object hypotheses thus differs from the selective processing of spatial locations.

Adaptive methods further determine where additional detail or computation is allocated. AdaBEV~\cite{li2025adabev} refines the foreground BEV cells identified through auxiliary 2D detections for collaborative drone perception. It then pools the refined subcell features into their parent cells while background queries are retained. A cooperative perception method~\cite{yazgan2026dense} maintains coarse coverage of the scene and allocates a communication budget to fine patches according to their expected benefit for detection. Task relevance also extends beyond foreground detection. SToRe3D~\cite{papais2026store3d} prunes image tokens and object queries with planning aligned relevance labels derived from the future interaction of the ego vehicle with other agents. BEVCast differs in the refinement criterion and in the representation handed to the trajectory head. Its heat head learns from the recorded path rather than from foreground membership or detection benefit, and its refined cells remain separate queries next to the unrefined coarse cells instead of being pooled back into their parents.

\subsection{Trajectory guided scene encoding}
\label{sec:related_trajectory_guided}

Predicted ego motion can guide both the readout of an existing scene representation and the extraction of image features. ThinkTwice~\cite{jia2023thinktwice} and VeteranAD~\cite{zhang2026veteranad} and SafeDrive~\cite{kim2026safedrive} query a scene BEV around the predicted ego position or along candidate trajectories, and SpatialBEV~\cite{nye2026structured} queries a terrain BEV at the tire tracks of predicted states for off road dynamics prediction. This readout operates after the BEV is built, so the cost of building that representation has already been paid. Other methods use trajectory proposals to organize the extraction of image features directly. iPad~\cite{guo2025ipad} anchors its feature sampling on candidate trajectories and refines the proposals and their features iteratively before it scores them. HiPro-AD~\cite{chen2026hiproad} extends proposal anchored attention with historical context, while S2Planner~\cite{lu2026s2planner} samples the features of three front cameras at the projected waypoints of its candidates. iPad and S2Planner represent nothing outside the candidate paths and both take a navigation command.

These approaches establish trajectories as a useful guide for selecting scene features. BEVCast applies this guidance to the spatial resolution of the BEV itself. The heat predicted for every coarse cell determines where fine queries are allocated, while the rest of the scene retains its coarse coverage in a mixed resolution BEV. The allocation thus takes place within the construction of the BEV and before the trajectory is decoded, which concentrates spatial detail where ego trajectory prediction needs it and reduces the BEV processing. The recorded path supervises the heat head during training only and inference uses the observed images and the ego motion without a navigation command.

\section{Method}
\label{sec:method}

\begin{figure*}[t]
    \centering
    \includegraphics[width=\textwidth]{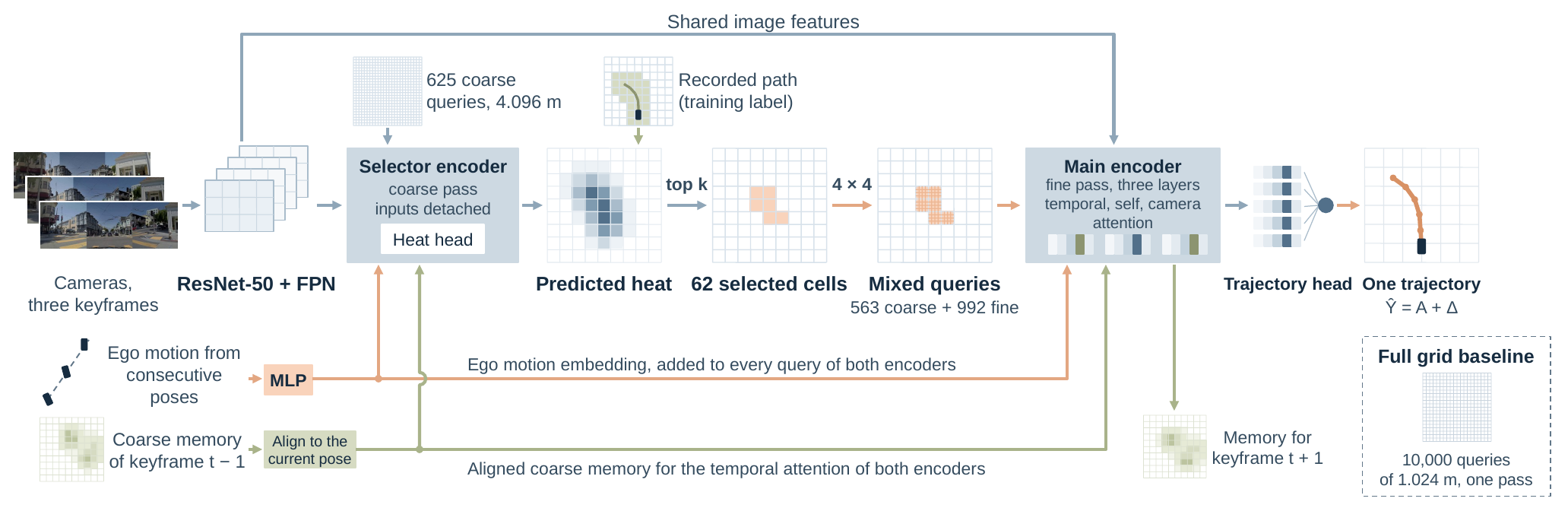}
    \caption{Overview of BEVCast at the 10\% refinement budget. The coarse pass encodes 625 cells of 4.096 m over the whole scene and its heat head predicts where the ego vehicle will drive. The 62 cells with the highest heat are subdivided into 1.024 m cells and the fine pass encodes these 992 fine cells together with the 563 unrefined coarse cells, from which the trajectory head regresses one trajectory. Both encoders read the shared image features and the ego motion embedding and the coarse memory of the previous keyframe. The full grid baseline encodes 10{,}000 cells of 1.024 m in one pass with the same encoder architecture.}
    \label{fig:overview}
\end{figure*}

\subsection{Overview}
\label{sec:setting}
Figure~\ref{fig:overview} shows the pipeline with its coarse pass (Section~\ref{sec:coarse}) and fine pass (Section~\ref{sec:refine}) and the memory pass that links the keyframes (Section~\ref{sec:memory}). At keyframe $t_0$ the model receives the images of $N_{\mathrm{cam}}$ cameras at $t_0$ and at up to two preceding keyframes with their projection matrices, together with the ego motion derived from the preceding poses. It outputs the positions $\hat{Y}\in\mathbb{R}^{T\times 2}$ of the rear axle at the next $T$ keyframes in the ego frame of $t_0$, with $T=6$ keyframes over 3 s on nuScenes and $T=8$ keyframes over 4 s on NAVSIM. Every BEV in this paper covers a square of side 102.4 m centred on the ego vehicle.

\subsection{BEV encoder on a set of cells}
\label{sec:encoder}
A BEV query is a feature token attached to a square cell with centre $(x_i,y_i)$ and side $c_i$. The encoder takes an arbitrary set of such cells rather than a grid and otherwise follows BEVFormer~\cite{li2022bevformer}. The input query of cell $i$ is
\begin{equation}
\mathbf{q}_i^{(0)}=\phi_{\mathrm{pos}}(x_i,y_i)+\phi_{\mathrm{size}}(c_i)+\mathbf{e}_{\mathrm{ego}},
\label{eq:bev_query}
\end{equation}
where $\phi_{\mathrm{pos}}$ maps Fourier features of the normalised centre with 16 frequencies through an MLP and $\phi_{\mathrm{size}}$ is a learned embedding of the cell size and $\mathbf{e}_{\mathrm{ego}}$ is the ego motion embedding of Section~\ref{sec:head}. Each of the three encoder layers first lets every query read the memory of the previous keyframe through temporal attention (Section~\ref{sec:memory}) and then applies self attention over the queries and deformable cross attention~\cite{zhu2021deformable} to the image features and a feed forward network. For the cross attention every cell lifts four points at heights from 4 m below to 2 m above the ego origin and projects them into every camera that sees them. It samples two learned offsets around each projection on each of the four FPN levels and averages the features over those cameras. The self attention costs $O(N_q^2)$ and the cross attention $O(N_q)$ in the number of queries $N_q$, so the query count is the natural measure of the encoder cost and we report it together with the measured time.

\subsection{Coarse pass and heat head}
\label{sec:coarse}
The coarse grid $G_c$ has $25\times 25$ cells of side $c_c=4.096$ m. A separate encoder with the architecture of Section~\ref{sec:encoder} encodes $G_c$ from the image features and the ego embedding and the coarse memory. All three inputs are detached, so this selector encoder is trained by its own loss alone. A two layer MLP called the heat head maps the encoded feature of every coarse cell $i$ to a logit $h_i$, and we call the score $s_i=\sigma(h_i)$ the heat of the cell. The heat target $u_i\in\{0,1\}$ marks the coarse cells that the recorded path crosses between $t_0$ and the last target keyframe, dilated by one cell in every direction. On nuScenes the recorded path is the sequence of ego poses of the 20 Hz LiDAR sweeps up to the last target keyframe and on NAVSIM it is the sequence of the future keyframe poses. The recorded path is used only to build these targets, so at inference the heat head predicts the heat from the observed images and the ego motion alone.

\subsection{Refinement and fine pass}
\label{sec:refine}
The refinement budget $b$ is the fraction of coarse cells that receive fine resolution. With $k=\operatorname{round}(b\,|G_c|)$ let $S_k\subset G_c$ be the $k$ cells with the highest heat. The selection is a hard top $k$ and passes no gradient. Each selected cell $i$ is subdivided into the set $\mathcal{F}(i)$ of its $4\times 4$ cells of side $c_f=1.024$ m. The query set of the fine pass keeps every unrefined coarse cell and replaces each selected cell by its fine cells. It is
\begin{equation}
Q=(G_c\setminus S_k)\cup\bigcup_{i\in S_k}\mathcal{F}(i)
\label{eq:query_set}
\end{equation}
and its $|Q|=|G_c|+15k$ cells tile the square without gaps at two resolutions. The main encoder encodes this mixed resolution BEV. Budgets of 5\% and 10\% and 20\% select 31 and 62 and 125 of the 625 coarse cells and change neither the algorithm nor the parameter count. At $b=0.1$ the set $Q$ contains 992 fine cells and 563 coarse cells, or 1{,}555 queries in total. Unless a budget is stated, BEVCast refers to this configuration.

A coarse cell contains $r\times r$ fine cells with $r$ dividing the 100 fine cells per side. With $r=2$ the coarse pass alone would need 2{,}500 queries and with $r=4$ it needs 625 or 6\% of the full grid, while a cell of 4.096 m is of the order of a lane width and a car length. On nuScenes the recorded 3 s path crosses 5.0 of these cells on average and at most 11. We did not sweep $r$.

\subsection{Temporal memory on the coarse grid}
\label{sec:memory}
With three keyframes the model follows the temporal self attention of BEVFormer. The keyframes are processed from the oldest to $t_0$ and every BEV query of the current keyframe attends to the BEV of the previous keyframe at its own position expressed in the previous ego frame, where it samples four learned offsets. The memory of BEVCast is the coarse grid, so the history of the whole scene persists across changes in the refined region. For every keyframe a memory pass of the main encoder encodes the 625 coarse cells while reading the aligned memory of the previous keyframe, and the result becomes the memory for the next keyframe. For a history keyframe this is all that is computed, without gradient. At $t_0$ the coarse pass and the fine pass both read this memory. The memory of the full grid baseline is the full grid itself, so that baseline is BEVFormer with a grid of 10{,}000 cells. Its ResNet-50 backbone and three encoder layers are those of the released tiny configuration of BEVFormer, which encodes 2{,}500 queries over the same square. At the 10\% budget BEVCast encodes $625+625+1{,}555=2{,}805$ queries per frame against $10{,}000$ for the full grid, and $4{,}055$ against $30{,}000$ over a training sample of three keyframes.

\subsection{Ego motion and trajectory head}
\label{sec:head}
The ego motion input has 24 numbers. Four are the forward and lateral speed and the forward acceleration and the yaw rate from the differences of consecutive keyframe poses. The other 20 are the position and heading and timestamp and a validity flag of each of the four preceding keyframes. An MLP with a final layer normalisation maps them to the 256 dimensional embedding $\mathbf{e}_{\mathrm{ego}}$ that Eq.~\ref{eq:bev_query} adds to every BEV query.

A two layer transformer decoder reads the encoded BEV with a single trajectory query. This query is the sum of a learned token and an MLP embedding of an anchor $A$, the mean future trajectory of the training set. Each decoder layer applies self attention and cross attention to the encoded BEV features and a feed forward network. A zero initialised linear layer regresses a residual $\Delta\in\mathbb{R}^{T\times 2}$ from the final trajectory token and the output is $\hat{Y}=A+\Delta$. The trajectory query receives no direct ego motion input and reads the ego state only through the BEV, because a planner with direct access to the ego state stops using the images under the nuScenes protocol~\cite{li2024egostatus} and a comparison between BEV layouts would then be void.

\subsection{Training objective}
\label{sec:training}
Training supervises the trajectory and the heat and the agent occupancy of the cells in $Q$. Let $\mathcal{V}$ be the set of valid target keyframes of a sample. The trajectory loss compares the predicted coordinates $(\hat{x}_t,\hat{y}_t)$ with the recorded coordinates $(x_t,y_t)$ in the ego frame of $t_0$,
\begin{equation}
\mathcal{L}_{\mathrm{traj}}=\frac{1}{|\mathcal{V}|}\sum_{t\in\mathcal{V}}\left[\rho(\hat{x}_t-x_t)+\rho(\hat{y}_t-y_t)\right],
\label{eq:traj_loss}
\end{equation}
where $\rho$ is the smooth L1 penalty with transition point $\beta=1$. For a logit $h$ and a binary target $u$ let
\begin{equation}
\ell_{\alpha}(h,u)=-\alpha\,u\log\sigma(h)-(1-u)\log\bigl(1-\sigma(h)\bigr)
\label{eq:bce}
\end{equation}
be the binary cross entropy whose weight $\alpha$ raises the cost of a missed positive cell. The heat loss trains the selector on the heat targets of Section~\ref{sec:coarse} with $\alpha=10$,
\begin{equation}
\mathcal{L}_{\mathrm{heat}}=\frac{1}{|G_c|}\sum_{i\in G_c}\ell_{\alpha}(h_i,u_i).
\label{eq:heat_loss}
\end{equation}
An agent occupancy head predicts a logit $o_j$ from the encoded feature of every cell $j\in Q$. Its target $v_j$ is 1 when the centre of the cell lies inside an agent box annotated at $t_0$ and 0 otherwise, and the occupancy loss with $\alpha=5$ is
\begin{equation}
\mathcal{L}_{\mathrm{occ}}=\frac{1}{|Q|}\sum_{j\in Q}\ell_{\alpha}(o_j,v_j).
\label{eq:occ_loss}
\end{equation}
The total loss is $\mathcal{L}=\mathcal{L}_{\mathrm{traj}}+\mathcal{L}_{\mathrm{heat}}+\mathcal{L}_{\mathrm{occ}}$ with unit weights and every term is averaged over the minibatch. The heat loss updates only the selector encoder and the heat head because the selector receives detached inputs. The occupancy head is attached to every model and the full grid baseline has no selector and no heat loss.

\section{Experiments}
\label{sec:experiments}

\subsection{Experimental setup}
\label{sec:setup}
All models share the backbone and the main encoder architecture and the trajectory head and the training recipe and receive three keyframes. The \emph{full grid} baseline encodes a uniform grid of $100\times 100$ cells of 1.024 m in one pass and is the reference of every ablation. The \emph{coarse only} model encodes the 625 cells of 4.096 m in one pass without selection and refinement and isolates what the learned refinement adds. The \emph{refined cells only} variant drops the 563 unrefined coarse cells as proposal centric planners do. The \emph{random selection} variant refines 62 coarse cells drawn at random in training and at inference and so needs no coarse pass. BEVCast is trained at refinement budgets of 5\% and 10\% and 20\%. We adopted 10\% as the common configuration after inspecting both benchmarks and report the other budgets next to it.

On nuScenes~\cite{caesar2020nuscenes} we use the official split of 700 training and 150 validation scenes with the annotated keyframes at 2 Hz. A sample needs a preceding keyframe for the ego motion and six future keyframes as targets, which leaves 23{,}230 training frames and 4{,}969 validation frames. The six images of 1600 by 900 pixels are resized to 800 by 448. No map and no CAN bus signal enters the model and the drivable area polygons of the map expansion are used only for the violation metric.

The evaluation follows the open loop protocol of ST-P3~\cite{hu2022stp3} as used by UniAD~\cite{hu2023planning} and VAD~\cite{jiang2023vad}. The L2 error at 1 s and 2 s and 3 s is the displacement at the keyframe nearest to that time and the mean of the three is the headline number. UniAD takes these values at the keyframe of each time, while ST-P3 and VAD and most later planners average them over all keyframes up to each time~\cite{weng2024paradrive}. Our ablations use the first convention and Table~\ref{tab:literature} the second. The error is also split into its longitudinal and lateral components along and across the recorded direction of travel. The collision rate at each time counts the frames in which the ego box of 1.85 m by 4.084 m placed 1.465 m ahead of the predicted rear axle overlaps a vehicle or pedestrian box annotated at that keyframe, counting only agents present at $t_0$ as in ST-P3 and VAD. The drivable area violation rate of BEV-Planner~\cite{li2024egostatus} counts the frames in which a corner of the ego box oriented along the predicted trajectory lies outside the drivable area of the map expansion. Steps where the recorded trajectory itself collides or leaves the drivable area are excluded from these rates. We also report the error at 3 s on the hard frames where a constant turn rate and acceleration extrapolation misses the recorded position by more than 2 m.

On NAVSIM we use the v2 devkit~\cite{dauner2024navsim,cao2025pseudosim} on OpenScene logs. The models train on 1{,}167 navtrain logs with 100{,}413 frames and are evaluated on the navtest split with 136 logs and 12{,}146 frames. The eight images of 1920 by 1080 pixels are resized to 800 by 448. The horizon is 4 s with eight target poses at 2 Hz and the recorded path for the heat target is the sequence of those poses. The ego motion is derived from consecutive poses as on nuScenes and the logged velocity and the driving command are not used. The model outputs positions and the heading required by the simulator is the central difference of consecutive predicted positions. The EPDMS is computed with the official devkit in the nonreactive navtest setting from one metric cache for every model. Recent planners report their scores on the same split~\cite{yao2025drivesuprim,li2025ztrs,wozniak2025prix,song2025diver,zou2025diffusiondrivev2}. The devkit sets a sub score to one when the recorded driver fails the same test in the same frame. The published navtest scores come from a devkit version that applied this rule to the printed sub scores and not to the total, so we report the sub scores with the rule and the total without it. Under this convention the recorded trajectory replayed through our pipeline scores 90.25 against the 90.3 reported for the human agent~\cite{yao2025drivesuprim}.

The backbone is a ResNet-50~\cite{he2016resnet} from ImageNet weights with its stem and first two stages frozen and an FPN~\cite{lin2017fpn} of four levels at strides 8 to 64 with 256 channels. All transformer layers have width 256 and eight heads, the encoder has three layers and the trajectory decoder two. We train for eight epochs with AdamW~\cite{loshchilov2019adamw} at a learning rate of 0.0002 for the BEV part and one tenth of it for the backbone and the FPN. Weight decay is 0.01, gradient clipping 5 and the learning rate follows a cosine schedule down to 5\% of its peak after 500 warm up steps. Batches hold eight frames in bfloat16 mixed precision. Every reported number comes from the model after the eighth epoch.

Cost is reported as the number of BEV queries per frame and as the measured time of the BEV part of the model, which is every encoder pass and the trajectory head. It is timed on precomputed backbone features in bfloat16 on an idle NVIDIA RTX PRO 6000 at batch size 1 as the median over 100 validation frames.

\begin{table*}[t]
\centering
\scriptsize
\setlength{\tabcolsep}{3.6pt}
\newcommand{\twohead}[1]{\multirow{2}{*}[-0.4ex]{#1}}
\begin{tabular}{l c c c c c c c | c c c c | c c c c}
\toprule
\twohead{\textbf{nuScenes planners}} & \twohead{Backbone} & \twohead{Sensors} & \twohead{Planner} & \twohead{Cmd} & \multicolumn{2}{c}{Ego status\rule[-1.1ex]{0pt}{0pt}} & \twohead{Map} & \twohead{L2 1 s\lb} & \twohead{2 s\lb} & \twohead{3 s\lb} & \twohead{Mean\lb} & \twohead{Col 1 s\lb} & \twohead{2 s\lb} & \twohead{3 s\lb} & \twohead{Mean\lb} \\
\noalign{\vskip-\aboverulesep}\cmidrule(lr){6-7}\noalign{\vskip-\belowrulesep\vskip-\cmidrulewidth}
 & & & & & BEV\rule{0pt}{2.3ex} & Planner & & & & & & & & & \\
\midrule
AD-MLP~\cite{zhai2023admlp} & \nodata & ego state & regression & \yes & \no & \yes & \no & 0.20 & 0.26 & 0.41 & 0.29 & 0.17\% & 0.18\% & 0.24\% & 0.19\% \\
ST-P3~\cite{hu2022stp3} & EfficientNet-B4 & C & sampling & \yes & \no & \no & \yes & 1.33 & 2.11 & 2.90 & 2.11 & 0.23\% & 0.62\% & 1.27\% & 0.71\% \\
VAD-Base~\cite{jiang2023vad} & ResNet-50 & C & regression & \yes & \yes & \no & \yes & 0.41 & 0.70 & 1.05 & 0.72 & 0.07\% & 0.17\% & 0.41\% & 0.22\% \\
BEV-Planner~\cite{li2024egostatus} & ResNet-50 & C & regression & \yes & \no & \no & \no & 0.30 & 0.52 & 0.83 & 0.55 & 0.10\% & 0.37\% & 1.30\% & 0.59\% \\
PARA-Drive~\cite{weng2024paradrive} & ResNet-50 & C & regression & \yes & \no & \no & \yes & 0.25 & 0.46 & 0.74 & 0.48 & 0.14\% & 0.23\% & 0.39\% & 0.25\% \\
SparseDrive-B~\cite{sun2024sparsedrive} & ResNet-101 & C & multi mode & \yes & \no & \no & \yes & 0.29 & 0.55 & 0.91 & 0.58 & 0.01\% & 0.02\% & 0.13\% & 0.06\% \\
\textbf{BEVCast 10\% (ours)} & ResNet-50 & C & regression & \no & \yes & \no & \no & 0.24 & 0.51 & 0.90 & 0.55 & 0.01\% & 0.06\% & 0.30\% & 0.12\% \\
\bottomrule
\end{tabular}
\caption{Published nuScenes planners next to BEVCast with the L2 error and the collision rate averaged over all keyframes up to each time as in ST-P3 and VAD and Mean over the three times. Sensors are C for cameras and Planner the type of trajectory output. Cmd marks a navigation command derived from the recorded future. Ego status marks an ego state input to the BEV or to the planning head and BEVCast derives this input from consecutive poses. Map marks map annotations as input or supervision. The published rates use a 0.5 m raster except BEV-Planner and SparseDrive which use oriented boxes, and BEVCast uses exact boxes with pedestrians on its 4{,}969 frames. Arrows mark the better direction.}
\label{tab:literature}
\end{table*}

\begin{table*}[t]
\centering
\scriptsize
\setlength{\tabcolsep}{3.4pt}
\begin{tabular}{l c c c c | c c c c c c c c c | c}
\toprule
\textbf{NAVSIM navtest} & Backbone & Sensors & Planner & Map & NC\hb & DAC\hb & DDC\hb & TLC\hb & EP\hb & TTC\hb & LK\hb & HC\hb & EC\hb & EPDMS\hb \\
\midrule
TransFuser~\cite{chitta2023transfuser,dauner2024navsim} & ResNet-34 & C+L & regression & \yes & 96.9 & 89.9 & 97.8 & 99.7 & 87.1 & 95.4 & 92.7 & 98.3 & 87.2 & 76.7 \\
Hydra-MDP++~\cite{li2025hydrampp,yao2025drivesuprim} & ResNet-34 & C & vocabulary & \yes & 97.2 & 97.5 & 99.4 & 99.6 & 83.1 & 96.5 & 94.4 & 98.2 & 70.9 & 81.4 \\
DriveSuprim~\cite{yao2025drivesuprim} & ResNet-34 & C & vocabulary & \yes & 97.5 & 96.5 & 99.4 & 99.6 & 88.4 & 96.6 & 95.5 & 98.3 & 77.0 & 83.1 \\
\textbf{BEVCast 10\% (ours)} & ResNet-50 & C & regression & \no & 98.0 & 94.3 & 99.5 & 99.8 & 87.1 & 97.4 & 97.1 & 98.3 & 84.6 & 82.5 \\
\bottomrule
\end{tabular}
\caption{Published planners on NAVSIM v2 navtest next to BEVCast. Sensors are C for cameras and L for LiDAR, Planner the type of trajectory output and Map the use of map annotations as input or supervision. The published planners receive the ego status and the driving command of the devkit, while BEVCast derives its ego motion from consecutive poses and receives no command. The TransFuser and Hydra-MDP++ values are those reported by DriveSuprim~\cite{yao2025drivesuprim} and all scores follow the convention of Section~\ref{sec:setup}. Arrows mark the better direction.}
\label{tab:literature_navsim}
\end{table*}

\subsection{Comparison with published planners}
\label{sec:literature}
Table~\ref{tab:literature} places BEVCast next to published nuScenes planners that output one trajectory. The published planners receive a navigation command derived from the recorded future and most of them train with detection and map supervision. BEVCast receives no command and no map and uses the agent boxes only for its occupancy loss. BEVCast and VAD-Base feed an ego state to the BEV and not to the planning head, and the mean L2 of BEVCast is 0.55 m against 0.72 m for VAD-Base. BEV-Planner reaches the same 0.55 m and SparseDrive-B 0.58 m without an ego state input. The collision rate of BEVCast is 0.12\% and lies between SparseDrive-B at 0.06\% and VAD-Base at 0.22\%. At 1 s its error of 0.24 m is the lowest of the camera based rows and at 3 s its 0.90 m lies behind PARA-Drive and BEV-Planner, which is consistent with the missing command mattering more at longer horizons. AD-MLP reaches 0.29 m from the ego state and the command alone~\cite{zhai2023admlp,li2024egostatus}.

Table~\ref{tab:literature_navsim} places BEVCast next to published navtest scores. All published rows use a ResNet-34 backbone and receive the ego status and the driving command of the devkit. TransFuser and BEVCast regress one trajectory, while Hydra-MDP++ and DriveSuprim score a vocabulary of trajectories with heads trained on the metric. TransFuser fuses a LiDAR sweep with the images and the other rows use cameras only. The EPDMS of BEVCast is 82.5, which lies above TransFuser at 76.7 and Hydra-MDP++ at 81.4 and below DriveSuprim at 83.1. Its no at fault collision and time to collision scores of 98.0 and 97.4 are the highest of the table, while its drivable area compliance of 94.3 lies between TransFuser at 89.9 and the two vocabulary planners at 96.5 and 97.5. With larger backbones the vocabulary planners reach 85.1 to 87.1~\cite{yao2025drivesuprim,li2025ztrs}.

\begin{figure*}[t]
    \centering
    \includegraphics[width=0.88\textwidth]{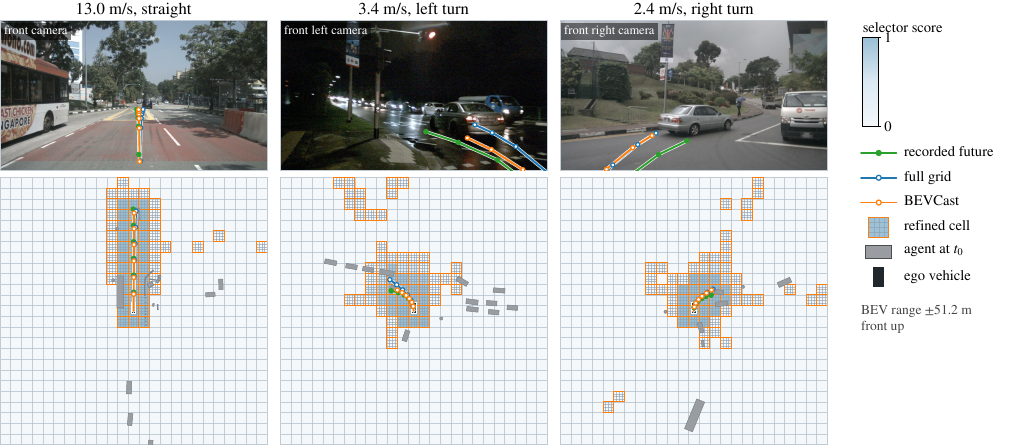}
    \caption{Learned selection at the 10\% refinement budget on three validation frames. The top row shows the camera that faces the recorded path with the recorded future in green and the predictions of the full grid in blue and of BEVCast in orange. The bottom row shows the heat as the selector score in pale blue on the coarse grid over the whole 102.4 m range with the 62 refined cells outlined in orange and subdivided into 1.024 m cells and the annotated agents in grey.}
    \label{fig:selection}
\end{figure*}

\subsection{Learned selection}
\label{sec:selection}
Figure~\ref{fig:selection} shows the refined cells on three validation frames. They form a corridor that extends past the end of the recorded path on the straight frame and cover the inside of the bend on the two turns. The figure shows where the fine resolution is spent rather than a consistent difference in accuracy, since the full grid is closer on the straight frame and BEVCast on the turns. Over the validation set the centre of the refined region lies 6.6 m ahead of the vehicle on frames slower than 3 m/s and 18.6 m ahead on frames faster than 8 m/s, and on frames with a yaw rate above 0.1 rad/s it shifts 2.0 m to the left on left turns and 2.2 m to the right on right turns. At the 10\% budget the refined region contains every coarse cell of the recorded path on all 4{,}969 frames and at 5\% it contains 99.9\% of them on average. Replacing the images of a frame by those of another scene changes the selection to a Jaccard overlap of 0.595 with the original and raises the 3 s endpoint error by 1.79 m, so the selection depends on the images as well as on the ego motion. The selection is learned from the recorded path with the same recipe on both benchmarks and needs no hand designed region.

\begin{table}[t]
\centering
\scriptsize
\setlength{\tabcolsep}{1.4pt}
\begin{tabular}{l r | c c c | c c c | c}
\toprule
\textbf{nuScenes 3 s} & Queries\lb & L2\lb & Lat\lb & Lon\lb & Col\lb & NM\lb & DAV\lb & Hard\lb \\
\midrule
Full grid 1.024 m & 10,000 & 1.149 & 0.206 & 1.083 & 0.44\% & 2.32\% & 2.29\% & 2.32 \\
Coarse only 4.096 m & 625 & 1.170 & 0.201 & 1.108 & 0.48\% & 2.41\% & 1.70\% & 2.32 \\
\midrule
BEVCast 20\% & 3,750 & 1.069 & 0.207 & 1.002 & 0.42\% & 2.26\% & 2.21\% & 2.38 \\
BEVCast 10\% & 2,805 & 1.078 & 0.191 & 1.022 & 0.40\% & 2.08\% & 1.46\% & 2.32 \\
BEVCast 5\% & 2,340 & 1.047 & 0.175 & 0.996 & 0.37\% & 1.52\% & 1.25\% & 2.24 \\
Refined cells only 10\% & 2,242 & 1.069 & 0.186 & 1.012 & 0.47\% & 2.05\% & 1.60\% & 2.27 \\
Random selection 10\% & 2,180 & 1.160 & 0.206 & 1.095 & 0.53\% & 2.51\% & 2.03\% & 2.33 \\
\bottomrule
\end{tabular}
\caption{nuScenes ablation with the 3 s protocol on the 4{,}969 validation frames. Queries are the BEV queries encoded per frame. L2 is the mean error in metres at the keyframes of 1 s to 3 s with Lat and Lon its lateral and longitudinal components. Col is the collision rate, NM the near miss rate with the ego box enlarged by 1.0 m and DAV the drivable area violation rate, all averaged over the three times. Hard is the error at 3 s on the 1{,}859 hard frames. BEVCast 10\% is the main configuration.}
\label{tab:nuscenes}
\end{table}

\subsection{Ablation study}
\label{sec:ablation}
Table~\ref{tab:nuscenes} reports the 3 s protocol on the 4{,}969 validation frames. BEVCast encodes 2{,}805 queries per frame against 10{,}000 for the full grid. Its mean L2 is 1.078 m against 1.149 m and its lateral error is 0.191 m against 0.206 m. The drivable area violation rate falls from 2.29\% to 1.46\% and the collision rate is 0.40\% against 0.44\%.

The coarse only model reaches a mean L2 of 1.170 m and a violation rate of 1.70\%, so one pass over 625 coarse cells already retains much of the accuracy of 10{,}000 fine cells. The learned refinement then lowers the mean L2 to 1.078 m and the violation rate to 1.46\%, with a longitudinal gain that reaches 0.37 m on the 728 stationary frames. The random selection variant refines as many cells and reaches 1.160 m and 2.03\%, so the gain comes from where the fine cells are placed and not from their number.

All three budgets have a lower mean L2 and a lower violation rate than the full grid. The 5\% budget has the lowest nuScenes values with 1.047 m and 1.25\%. The 10\% budget gives 1.078 m and 1.46\% and the 20\% budget gives 1.069 m and 2.21\%, so more refinement does not improve these models.

Dropping the 563 unrefined coarse cells leaves the trajectory head with the 992 refined cells. This variant reaches a mean L2 of 1.069 m and a violation rate of 1.60\% against 1.078 m and 1.46\% for BEVCast, so the displacement accuracy is comparable and the coarse context lowers the violation rate. We did not run this variant on NAVSIM. Marking the history keyframes as invalid at inference raises the mean L2 by 0.43 m for BEVCast and by 0.42 m for this variant against 0.86 m for the full grid and 0.92 m for the coarse only model. The two layouts that refine a coarse pass thus depend on the memory about half as much as the two single resolution grids.

\begin{table}[t]
\centering
\scriptsize
\setlength{\tabcolsep}{0.9pt}
\begin{tabular}{l | c c c c c c c | c}
\toprule
\textbf{Violation rate} & Static\lb & Start\lb & Stop\lb & Left\lb & Right\lb & Turns\lb & Straight\lb & All\lb \\
\secondary{Frames} & \secondary{728} & \secondary{207} & \secondary{100} & \secondary{240} & \secondary{339} & \secondary{579} & \secondary{3,355} & \secondary{4,969} \\
\midrule
Full grid 1.024 m & 1.92\% & 3.06\% & 7.33\% & 5.14\% & 9.24\% & 7.54\% & 1.27\% & 2.29\% \\
Coarse only 4.096 m & 0.82\% & 4.03\% & 3.67\% & 3.89\% & 7.28\% & 5.87\% & 0.96\% & 1.70\% \\
\midrule
BEVCast 20\% & 1.83\% & 2.74\% & 3.33\% & 7.36\% & 8.46\% & 8.00\% & 1.22\% & 2.21\% \\
BEVCast 10\% & 0.73\% & 3.22\% & 4.67\% & 4.86\% & 4.72\% & 4.78\% & 0.83\% & 1.46\% \\
BEVCast 5\% & 0.60\% & 2.09\% & 3.33\% & 4.86\% & 4.72\% & 4.78\% & 0.67\% & 1.25\% \\
\bottomrule
\end{tabular}
\caption{Drivable area violation rate on nuScenes by the behaviour of the recorded future over the next 3 s. Static is a displacement below 1 m, start a speed below 1 m/s with a displacement above 2 m and stop a moving vehicle that advances less than 0.5 m in the last second. Left and right turns follow the rule of ST-P3 with a lateral offset of at least 2 m at 3 s and the first matching label is taken in this order. Turns joins both. BEVCast 10\% is the main configuration.}
\label{tab:behaviour}
\end{table}

Table~\ref{tab:behaviour} splits the drivable area violation rate by the behaviour of the recorded future. Against the full grid BEVCast lowers the rate on the 339 right turns from 9.24\% to 4.72\% and on all 579 turning frames from 7.54\% to 4.78\%. It also lowers the rate on the straight and the stationary frames while the two rates are close on the left turns. The lateral error on the turning frames is 0.481 m for BEVCast against 0.557 m for the full grid. The 5\% and 10\% budgets violate on different turning frames but on the same number of steps, 35 of 720 on the left turns and 48 of 1{,}017 on the right turns.

\begin{table}[t]
\centering
\scriptsize
\setlength{\tabcolsep}{0.9pt}
\begin{tabular}{l | c c | c | c c c}
\toprule
\textbf{NAVSIM navtest} & L2 1 to 3 s\lb & 1 to 4 s\lb & Col\lb & EPDMS\hb & DAC\hb & EC\hb \\
\midrule
Full grid 1.024 m & 0.598 & 0.907 & 0.09\% & 81.28 & 93.4 & 83.6 \\
Coarse only 4.096 m & 0.585 & 0.887 & 0.07\% & 81.85 & 94.0 & 84.0 \\
\midrule
BEVCast 20\% & 0.590 & 0.894 & 0.05\% & 81.64 & 93.5 & 83.9 \\
BEVCast 10\% & 0.577 & 0.876 & 0.04\% & 82.51 & 94.3 & 84.6 \\
BEVCast 5\% & 0.580 & 0.881 & 0.06\% & 82.22 & 94.0 & 84.1 \\
\bottomrule
\end{tabular}
\caption{NAVSIM v2 navtest ablation on the 12{,}146 frames with a 4 s horizon and the queries of Table~\ref{tab:nuscenes}. L2 is the mean error in metres over 1 s to 3 s and over 1 s to 4 s and Col the collision rate over 1 s to 3 s. EPDMS follows Section~\ref{sec:setup} and DAC and EC are its drivable area compliance and extended comfort sub scores. BEVCast 10\% is the main configuration.}
\label{tab:navsim}
\end{table}

Table~\ref{tab:navsim} reports navtest with a horizon of 4 s at the same query counts. The mean L2 of BEVCast over 1 s to 3 s is 0.577 m against 0.598 m for the full grid and over 1 s to 4 s it is 0.876 m against 0.907 m. The EPDMS rises from 81.28 to 82.51 and the collision rate is 0.04\% against 0.09\%. Its drivable area compliance and extended comfort rise to 94.3 and 84.6 from 93.4 and 83.6. Its no at fault collision score rises to 98.0 from 97.5 and its time to collision score to 97.4 from 96.9, while its ego progress is slightly lower at 87.1 against 87.3.

The coarse only model reaches a mean L2 of 0.585 m and an EPDMS of 81.85 against 0.577 m and 82.51 for BEVCast. The learned refinement thus raises the EPDMS by 0.66 points on top of the coarse grid and improves its drivable area compliance and extended comfort.

The 5\% budget reaches a mean L2 of 0.580 m and an EPDMS of 82.22 and the 20\% budget 0.590 m and 81.64, so the additional 34\% of queries at 20\% do not improve these models.

\begin{table}[t]
\centering
\scriptsize
\setlength{\tabcolsep}{2.0pt}
\begin{tabular}{l r r r r r r}
\toprule
 & \multicolumn{2}{c}{Queries\lb} & \multicolumn{3}{c}{Time per sample (ms)\lb} & BEV\lb \\
\cmidrule(lr){2-3} \cmidrule(lr){4-6}
\textbf{Batch size 1} & frame & sample & Backbone & BEV & Model & vs full \\
\midrule
Full grid 1.024 m & 10,000 & 30,000 & 33.4 & 23.6 & 57.0 & 100\% \\
Coarse only 4.096 m & 625 & 1,875 & 33.3 & 9.5 & 42.8 & 40\% \\
\midrule
BEVCast 20\% & 3,750 & 5,000 & 33.3 & 15.7 & 49.1 & 67\% \\
BEVCast 10\% & 2,805 & 4,055 & 33.4 & 15.6 & 48.9 & 66\% \\
BEVCast 5\% & 2,340 & 3,590 & 33.4 & 15.5 & 48.9 & 66\% \\
Refined cells only 10\% & 2,242 & 3,492 & 33.4 & 15.4 & 48.8 & 65\% \\
\bottomrule
\end{tabular}
\caption{Measured cost on nuScenes frames at batch size 1 on an idle RTX PRO 6000 in bfloat16 with three keyframes per sample. Queries per sample include the two history keyframes, Model is the sum of the backbone and the BEV part and the last column is the BEV time relative to the full grid. Lower is better in every column. BEVCast 10\% is the main configuration.}
\label{tab:compute}
\end{table}

\subsection{Computational cost}
\label{sec:compute}
Table~\ref{tab:compute} gives the measured cost and efficiency at batch size 1. The BEV part of BEVCast runs in 15.6 ms per sample against 23.6 ms for the full grid, or 66\% of its time. The image backbone is shared by all layouts and takes 33 ms per sample of three keyframes, so the whole model runs in 48.9 ms against 57.0 ms or 86\%. The saving in time is smaller than the saving in queries because the fixed cost of an encoder pass dominates at batch size 1 and BEVCast runs three passes per frame against one for the full grid, namely the memory pass and the coarse pass and the fine pass. For the same reason the budgets from 5\% to 20\% lie within 0.3 ms of each other.

\section{Conclusion}
\label{sec:conclusion}

The spatial resolution of a driving BEV need not be uniform, since a coarse grid of 625 cells with fine cells on 62 of them predicts the ego trajectory better than the full grid of 10{,}000 fine cells on both benchmarks and keeps it on the road more often on nuScenes. The recorded path is a supervision signal for allocating computation that every driving log provides at no cost, and a selector trained on it covers every path cell of the validation frames and shifts with the speed and the turning direction. Finally the path adaptive allocation improves the trade off between accuracy and computation, since BEVCast exceeds the accuracy of the full grid with 28\% of the BEV queries and 66\% of the BEV encoder time.

The gains over the full grid belong to the two stage design as a whole rather than to the query count alone, since BEVCast also adds a selector encoder and a coarse memory grid. The single output trajectory falls between turning and going straight where both are plausible, which is the cost of working without a navigation command. The open loop L2 of nuScenes depends strongly on the ego state, so we report it together with the lateral error and the drivable area violations and the EPDMS of NAVSIM. The comparison with published planners is a reference for the magnitude and not a ranking, since their training settings and protocols differ from ours. Diffusion based trajectory heads improve on the planners in our tables on both benchmarks by changing the decoder on top of a scene representation. Pairing such a head with our BEV is left to future work together with closed loop replanning, which neither benchmark tests.

BEVCast builds a path adaptive BEV in two stages without a map or a navigation command. The resolution of a driving BEV can follow the task rather than the grid.

{
    \small
    \bibliographystyle{ieeenat_fullname}
    \bibliography{main}

\begin{thebibliography}{37}
\providecommand{\natexlab}[1]{#1}
\providecommand{\url}[1]{\texttt{#1}}
\expandafter\ifx\csname urlstyle\endcsname\relax
  \providecommand{\doi}[1]{doi: #1}\else
  \providecommand{\doi}{doi: \begingroup \urlstyle{rm}\Url}\fi

\bibitem[Caesar et~al.(2020)Caesar, Bankiti, Lang, Vora, Liong, Xu, Krishnan,
  Pan, Baldan, and Beijbom]{caesar2020nuscenes}
Holger Caesar, Varun Bankiti, Alex~H Lang, Sourabh Vora, Venice~Erin Liong,
  Qiang Xu, Anush Krishnan, Yu Pan, Giancarlo Baldan, and Oscar Beijbom.
\newblock {nuScenes}: A multimodal dataset for autonomous driving.
\newblock In \emph{Proceedings of the IEEE/CVF Conference on Computer Vision
  and Pattern Recognition}, pages 11621--11631, 2020.

\bibitem[Cao et~al.(2025)Cao, Hallgarten, Li, Dauner, Gu, Wang, Miron, Aiello,
  Li, Gilitschenski, Ivanovic, Pavone, Geiger, and Chitta]{cao2025pseudosim}
Wei Cao, Marcel Hallgarten, Tianyu Li, Daniel Dauner, Xunjiang Gu, Caojun Wang,
  Yakov Miron, Marco Aiello, Hongyang Li, Igor Gilitschenski, Boris Ivanovic,
  Marco Pavone, Andreas Geiger, and Kashyap Chitta.
\newblock Pseudo-simulation for autonomous driving.
\newblock In \emph{Conference on Robot Learning}, 2025.

\bibitem[Chambon et~al.(2024)Chambon, Zablocki, Chen, Bartoccioni, P{\'e}rez,
  and Cord]{chambon2024pointbev}
Loick Chambon, {\'E}loi Zablocki, Micka{\"e}l Chen, Florent Bartoccioni,
  Patrick P{\'e}rez, and Matthieu Cord.
\newblock {PointBeV}: A sparse approach for {BeV} predictions.
\newblock In \emph{Proceedings of the IEEE/CVF Conference on Computer Vision
  and Pattern Recognition}, pages 15195--15204, 2024.

\bibitem[Chen et~al.(2026)Chen, Wang, Yang, Huang, Qian, Huang, and
  Guo]{chen2026hiproad}
Bing Chen, Gaopeng Wang, Jiandong Yang, Shaoliang Huang, Xinhe Qian, Bin Huang,
  and Guanlun Guo.
\newblock {HiPro-AD}: Sparse trajectory transformer for end-to-end autonomous
  driving with hybrid spatiotemporal attention.
\newblock \emph{Sensors}, 26\penalty0 (1):\penalty0 185, 2026.

\bibitem[Chen et~al.(2024)Chen, Xu, Ye, Qian, Zou, Yeung, and
  Chen]{chen2024vectorformer}
Zhili Chen, Shuangjie Xu, Maosheng Ye, Zian Qian, Xiaoyi Zou, Dit-Yan Yeung,
  and Qifeng Chen.
\newblock Learning high-resolution vector representation from multi-camera
  images for {3D} object detection.
\newblock In \emph{European Conference on Computer Vision}, 2024.

\bibitem[Chitta et~al.(2023)Chitta, Prakash, Jaeger, Yu, Renz, and
  Geiger]{chitta2023transfuser}
Kashyap Chitta, Aditya Prakash, Bernhard Jaeger, Zehao Yu, Katrin Renz, and
  Andreas Geiger.
\newblock {TransFuser}: Imitation with transformer-based sensor fusion for
  autonomous driving.
\newblock \emph{IEEE Transactions on Pattern Analysis and Machine
  Intelligence}, 45\penalty0 (11):\penalty0 12878--12895, 2023.

\bibitem[Dauner et~al.(2024)Dauner, Hallgarten, Li, Weng, Huang, Yang, Li,
  Gilitschenski, Ivanovic, Pavone, Geiger, and Chitta]{dauner2024navsim}
Daniel Dauner, Marcel Hallgarten, Tianyu Li, Xinshuo Weng, Zhiyu Huang, Zetong
  Yang, Hongyang Li, Igor Gilitschenski, Boris Ivanovic, Marco Pavone, Andreas
  Geiger, and Kashyap Chitta.
\newblock {NAVSIM}: Data-driven non-reactive autonomous vehicle simulation and
  benchmarking.
\newblock In \emph{Advances in Neural Information Processing Systems}, 2024.

\bibitem[Guo et~al.(2026)Guo, Liu, Wu, Pan, and Lv]{guo2025ipad}
Ke Guo, Haochen Liu, Xiaojun Wu, Jia Pan, and Chen Lv.
\newblock {iPad}: Iterative proposal-centric end-to-end autonomous driving.
\newblock \emph{IEEE Robotics and Automation Letters}, 11\penalty0
  (10):\penalty0 11142--11149, 2026.

\bibitem[He et~al.(2016)He, Zhang, Ren, and Sun]{he2016resnet}
Kaiming He, Xiangyu Zhang, Shaoqing Ren, and Jian Sun.
\newblock Deep residual learning for image recognition.
\newblock In \emph{Proceedings of the IEEE Conference on Computer Vision and
  Pattern Recognition}, pages 770--778, 2016.

\bibitem[Hu et~al.(2022)Hu, Chen, Wu, Li, Yan, and Tao]{hu2022stp3}
Shengchao Hu, Li Chen, Penghao Wu, Hongyang Li, Junchi Yan, and Dacheng Tao.
\newblock {ST-P3}: End-to-end vision-based autonomous driving via
  spatial-temporal feature learning.
\newblock In \emph{European Conference on Computer Vision}, pages 533--549,
  2022.

\bibitem[Hu et~al.(2023)Hu, Yang, Chen, Li, Sima, Zhu, Chai, Du, Lin, Wang, Lu,
  Jia, Liu, Dai, Qiao, and Li]{hu2023planning}
Yihan Hu, Jiazhi Yang, Li Chen, Keyu Li, Chonghao Sima, Xizhou Zhu, Siqi Chai,
  Senyao Du, Tianwei Lin, Wenhai Wang, Lewei Lu, Xiaosong Jia, Qiang Liu,
  Jifeng Dai, Yu Qiao, and Hongyang Li.
\newblock Planning-oriented autonomous driving.
\newblock In \emph{Proceedings of the IEEE/CVF Conference on Computer Vision
  and Pattern Recognition}, pages 17853--17862, 2023.

\bibitem[Jia et~al.(2023)Jia, Wu, Chen, Xie, He, Yan, and
  Li]{jia2023thinktwice}
Xiaosong Jia, Penghao Wu, Li Chen, Jiangwei Xie, Conghui He, Junchi Yan, and
  Hongyang Li.
\newblock Think twice before driving: Towards scalable decoders for end-to-end
  autonomous driving.
\newblock In \emph{Proceedings of the IEEE/CVF Conference on Computer Vision
  and Pattern Recognition}, 2023.

\bibitem[Jiang et~al.(2023)Jiang, Chen, Xu, Liao, Chen, Zhou, Zhang, Liu,
  Huang, and Wang]{jiang2023vad}
Bo Jiang, Shaoyu Chen, Qing Xu, Bencheng Liao, Jiajie Chen, Helong Zhou, Qian
  Zhang, Wenyu Liu, Chang Huang, and Xinggang Wang.
\newblock {VAD}: Vectorized scene representation for efficient autonomous
  driving.
\newblock In \emph{Proceedings of the IEEE/CVF International Conference on
  Computer Vision}, 2023.

\bibitem[Kim et~al.(2026)Kim, Oh, Yu, Shin, Kwak, and Choi]{kim2026safedrive}
Jungho Kim, Jiyong Oh, Seunghoon Yu, Hongjae Shin, Donghyuk Kwak, and Jun~Won
  Choi.
\newblock {SafeDrive}: Fine-grained safety reasoning for end-to-end driving in
  a sparse world.
\newblock In \emph{Proceedings of the IEEE/CVF Conference on Computer Vision
  and Pattern Recognition}, 2026.

\bibitem[Li et~al.(2025{\natexlab{a}})Li, Li, Lan, Xie, Zhang, Liu, Wu, Yu, and
  Alvarez]{li2025hydrampp}
Kailin Li, Zhenxin Li, Shiyi Lan, Yuan Xie, Zhizhong Zhang, Jiayi Liu, Zuxuan
  Wu, Zhiding Yu, and Jose~M. Alvarez.
\newblock {Hydra-MDP++}: Advancing end-to-end driving via expert-guided
  {Hydra-Distillation}.
\newblock \emph{arXiv preprint arXiv:2503.12820}, 2025{\natexlab{a}}.

\bibitem[Li and Cui(2025)]{li2025ssr}
Peidong Li and Dixiao Cui.
\newblock Navigation-guided sparse scene representation for end-to-end
  autonomous driving.
\newblock In \emph{International Conference on Learning Representations}, 2025.

\bibitem[Li et~al.(2022)Li, Wang, Li, Xie, Sima, Lu, Qiao, and
  Dai]{li2022bevformer}
Zhiqi Li, Wenhai Wang, Hongyang Li, Enze Xie, Chonghao Sima, Tong Lu, Yu Qiao,
  and Jifeng Dai.
\newblock {BEVFormer}: Learning bird's-eye-view representation from
  multi-camera images via spatiotemporal transformers.
\newblock In \emph{European Conference on Computer Vision}, pages 1--18, 2022.

\bibitem[Li et~al.(2024)Li, Yu, Lan, Li, Kautz, Lu, and
  Alvarez]{li2024egostatus}
Zhiqi Li, Zhiding Yu, Shiyi Lan, Jiahan Li, Jan Kautz, Tong Lu, and Jose~M.
  Alvarez.
\newblock Is ego status all you need for open-loop end-to-end autonomous
  driving?
\newblock In \emph{Proceedings of the IEEE/CVF Conference on Computer Vision
  and Pattern Recognition}, 2024.

\bibitem[Li et~al.(2025{\natexlab{b}})Li, Cheng, Zhao, Chen, Li, Wang, Yang,
  Sun, and Wang]{li2025adabev}
Zhongyao Li, Peirui Cheng, Liangjin Zhao, Chen Chen, Yundu Li, Zhechao Wang,
  Xue Yang, Xian Sun, and Zhirui Wang.
\newblock Refine-and-contrast: Adaptive instance-aware {BEV} representations
  for multi-{UAV} collaborative object detection.
\newblock \emph{arXiv preprint arXiv:2508.12684}, 2025{\natexlab{b}}.

\bibitem[Li et~al.(2026)Li, Chang, Yao, Sun, Wang, Shen, Chen, Song, Li, Wu,
  Lan, and Alvarez]{li2025ztrs}
Zhenxin Li, Nadine Chang, Wenhao Yao, Xinglong Sun, Zi Wang, Maying Shen,
  Jingde Chen, Jingyu Song, Kailin Li, Zuxuan Wu, Shiyi Lan, and Jose~M.
  Alvarez.
\newblock Zero-human demonstration end-to-end autonomous driving with
  trajectory scorer.
\newblock In \emph{European Conference on Computer Vision}, pages 91--110,
  2026.

\bibitem[Liao et~al.(2025)Liao, Chen, Yin, Jiang, Wang, Yan, Zhang, Li, Zhang,
  Zhang, and Wang]{liao2025diffusiondrive}
Bencheng Liao, Shaoyu Chen, Haoran Yin, Bo Jiang, Cheng Wang, Sixu Yan, Xinbang
  Zhang, Xiangyu Li, Ying Zhang, Qian Zhang, and Xinggang Wang.
\newblock {DiffusionDrive}: Truncated diffusion model for end-to-end autonomous
  driving.
\newblock In \emph{Proceedings of the IEEE/CVF Conference on Computer Vision
  and Pattern Recognition}, 2025.

\bibitem[Lin et~al.(2017)Lin, Doll{\'a}r, Girshick, He, Hariharan, and
  Belongie]{lin2017fpn}
Tsung-Yi Lin, Piotr Doll{\'a}r, Ross Girshick, Kaiming He, Bharath Hariharan,
  and Serge Belongie.
\newblock Feature pyramid networks for object detection.
\newblock In \emph{Proceedings of the IEEE Conference on Computer Vision and
  Pattern Recognition}, pages 2117--2125, 2017.

\bibitem[Liu et~al.(2023)Liu, Teng, Lu, Wang, and Wang]{liu2023sparsebev}
Haisong Liu, Yao Teng, Tao Lu, Haiguang Wang, and Limin Wang.
\newblock {SparseBEV}: High-performance sparse {3D} object detection from
  multi-camera videos.
\newblock In \emph{Proceedings of the IEEE/CVF International Conference on
  Computer Vision}, 2023.

\bibitem[Loshchilov and Hutter(2019)]{loshchilov2019adamw}
Ilya Loshchilov and Frank Hutter.
\newblock Decoupled weight decay regularization.
\newblock In \emph{International Conference on Learning Representations}, 2019.

\bibitem[Lu et~al.(2026)Lu, Zhou, Guo, Yu, and Knoll]{lu2026s2planner}
Zhaowei Lu, Liguo Zhou, Yujie Guo, Lei Yu, and Alois Knoll.
\newblock {S2Planner}: Multi-scale semantic planner for end-to-end autonomous
  driving.
\newblock \emph{arXiv preprint arXiv:2609.29813}, 2026.

\bibitem[Nye et~al.(2026)Nye, Triest, Sivaprakasam, Adu, Scherer, and
  Wang]{nye2026structured}
Micah Nye, Samuel Triest, Matthew Sivaprakasam, Isaiah Adu, Sebastian Scherer,
  and Wenshan Wang.
\newblock Structured perception representations for off-road dynamics learning.
\newblock In \emph{Conference on Robot Learning}, 2026.

\bibitem[Papais et~al.(2026)Papais, Feng, Cossette, and Ge]{papais2026store3d}
Sandro Papais, Lezhou Feng, Charles Cossette, and Lingting Ge.
\newblock {SToRe3D}: Sparse token relevance in {ViTs} for efficient multi-view
  {3D} object detection.
\newblock \emph{arXiv preprint arXiv:2605.14110}, 2026.

\bibitem[Song et~al.(2026)Song, Liu, Pan, Liao, Guo, Yang, Zhang, Xu, Jia, and
  Luo]{song2025diver}
Ziying Song, Lin Liu, Hongyu Pan, Bencheng Liao, Mingzhe Guo, Lei Yang,
  Yongchang Zhang, Shaoqing Xu, Caiyan Jia, and Yadan Luo.
\newblock {DIVER}: Reinforced diffusion breaks imitation bottlenecks in
  end-to-end autonomous driving.
\newblock \emph{IEEE Transactions on Pattern Analysis and Machine
  Intelligence}, 48\penalty0 (11):\penalty0 13856--13872, 2026.

\bibitem[Sun et~al.(2025)Sun, Lin, Shi, Zhang, Wu, and
  Zheng]{sun2024sparsedrive}
Wenchao Sun, Xuewu Lin, Yining Shi, Chuang Zhang, Haoran Wu, and Sifa Zheng.
\newblock {SparseDrive}: End-to-end autonomous driving via sparse scene
  representation.
\newblock In \emph{IEEE International Conference on Robotics and Automation},
  2025.

\bibitem[Weng et~al.(2024)Weng, Ivanovic, Wang, Wang, and
  Pavone]{weng2024paradrive}
Xinshuo Weng, Boris Ivanovic, Yan Wang, Yue Wang, and Marco Pavone.
\newblock {PARA-Drive}: Parallelized architecture for real-time autonomous
  driving.
\newblock In \emph{Proceedings of the IEEE/CVF Conference on Computer Vision
  and Pattern Recognition}, 2024.

\bibitem[Wozniak et~al.(2026)Wozniak, Liu, Cai, and Jensfelt]{wozniak2025prix}
Maciej Wozniak, Lianhang Liu, Yixi Cai, and Patric Jensfelt.
\newblock {PRIX}: Learning to plan from raw pixels for end-to-end autonomous
  driving.
\newblock \emph{IEEE Robotics and Automation Letters}, 11\penalty0
  (5):\penalty0 6400--6407, 2026.

\bibitem[Yao et~al.(2026)Yao, Li, Lan, Wang, Sun, Alvarez, and
  Wu]{yao2025drivesuprim}
Wenhao Yao, Zhenxin Li, Shiyi Lan, Zi Wang, Xinglong Sun, Jose~M. Alvarez, and
  Zuxuan Wu.
\newblock {DriveSuprim}: Towards precise trajectory selection for end-to-end
  planning.
\newblock In \emph{Proceedings of the AAAI Conference on Artificial
  Intelligence}, pages 11910--11918, 2026.

\bibitem[Yazgan et~al.(2026)Yazgan, M{\"u}ller, and
  Z{\"o}llner]{yazgan2026dense}
Melih Yazgan, Timon M{\"u}ller, and J.~Marius Z{\"o}llner.
\newblock Dense coverage, sparse refinement: Byte-constrained cooperative
  perception.
\newblock \emph{arXiv preprint arXiv:2609.29456}, 2026.

\bibitem[Zhai et~al.(2023)Zhai, Feng, Du, Mao, Liu, Tan, Zhang, Ye, and
  Wang]{zhai2023admlp}
Jiang-Tian Zhai, Ze Feng, Jinhao Du, Yongqiang Mao, Jiang-Jiang Liu, Zichang
  Tan, Yifu Zhang, Xiaoqing Ye, and Jingdong Wang.
\newblock Rethinking the open-loop evaluation of end-to-end autonomous driving
  in {nuScenes}.
\newblock \emph{arXiv preprint arXiv:2305.10430}, 2023.

\bibitem[Zhang et~al.(2026)Zhang, Li, Song, and Zhang]{zhang2026veteranad}
Bozhou Zhang, Jingyu Li, Nan Song, and Li Zhang.
\newblock Perception in plan: Coupled perception and planning for end-to-end
  autonomous driving.
\newblock In \emph{Proceedings of the AAAI Conference on Artificial
  Intelligence}, pages 12376--12384, 2026.

\bibitem[Zhu et~al.(2021)Zhu, Su, Lu, Li, Wang, and Dai]{zhu2021deformable}
Xizhou Zhu, Weijie Su, Lewei Lu, Bin Li, Xiaogang Wang, and Jifeng Dai.
\newblock Deformable {DETR}: Deformable transformers for end-to-end object
  detection.
\newblock In \emph{International Conference on Learning Representations}, 2021.

\bibitem[Zou et~al.(2025)Zou, Chen, Liao, Zheng, Song, Zhang, Zhang, Liu, and
  Wang]{zou2025diffusiondrivev2}
Jialv Zou, Shaoyu Chen, Bencheng Liao, Zhiyu Zheng, Yuehao Song, Lefei Zhang,
  Qian Zhang, Wenyu Liu, and Xinggang Wang.
\newblock {DiffusionDriveV2}: Reinforcement learning-constrained truncated
  diffusion modeling in end-to-end autonomous driving.
\newblock \emph{arXiv preprint arXiv:2512.07745}, 2025.

\end{thebibliography}
}

\end{document}